\documentclass[lettersize,journal]{IEEEtran}
\usepackage{amsmath,amsfonts}
\usepackage{algorithmic}
\usepackage{algorithm}
\usepackage{array}
\usepackage[caption=false,font=normalsize,labelfont=sf,textfont=sf]{subfig}
\usepackage{textcomp}
\usepackage{stfloats}
\usepackage{url}
\usepackage{verbatim}
\usepackage{graphicx}
\usepackage{cite}
\usepackage{multirow}
\usepackage{booktabs}   
\usepackage{makecell}   
\usepackage{tabularx}   
\usepackage{array}      
\newcolumntype{Y}{>{\centering\arraybackslash}X}
\usepackage{titling} 

\usepackage{float}
\usepackage[colorlinks=true,
            linkcolor=blue,
            urlcolor=blue,
            citecolor=blue]{hyperref}

\usepackage{enumitem}

\begin{document}

\title{REAP: Reinforcement-Learning End-to-End Autonomous Parking with Gaussian Splatting Simulator for Real2Sim2Real Transfer}

\author{Changze Li, Zhe Chen, Shaoyu Chen, Lisen Mu, Yijian Li, Yuelong Yu,  \\ Qian Zhang, Qing Su, Ming Yang, Tong Qin
\thanks{Changze Li, Zhe Chen, Yijian Li, Ming Yang and Tong Qin are with Shanghai Jiao Tong University, Shanghai, China.
Shaoyu Chen, Lisen Mu, Yuelong Yu, Qian Zhang and Qing Su are with Horizon Robotics Technology Research and Development Co., Ltd.
}
}

\renewcommand{\thefootnote}{}

\markboth{IEEE TRANSACTIONS ON INTELLIGENT TRANSPORTATION SYSTEMS}%
{Li \MakeLowercase{\textit{et al.}}: REAP: Reinforcement-Learning End-to-End Autonomous Parking for Real2Sim2Real Transfer}


\date{}
\maketitle

\begin{abstract}

In recent years, autonomous parking has made significant advances, yet parking tasks still face challenges in extreme scenarios such as mechanical and dead-end parking slots, often resulting in failures.
This is mainly due to traditional parking methods adopting a multistage approach, lacking the ability to optimize the parking problem as a whole.
End-to-end methods enable joint optimization across perception and planning modules to eliminate the accumulation of errors, enhancing algorithm performance in extreme scenarios.
Although several end-to-end parking methods use imitation or reinforcement learning, the former is limited by data cost and distribution coverage, while the latter suffers from inefficient exploration.
To address these challenges, we propose a \textbf{R}einforcement learning \textbf{E}nd-to-end \textbf{A}utonomous \textbf{P}arking method \textbf{(REAP)}. 
REAP employs Soft Actor-Critic (SAC) within an asymmetric reinforcement learning framework to improve training efficiency and inference performance.
To accelerate model convergence, we distill the capabilities of a rule-based planner into the end-to-end network through behavior cloning.
We further introduce a soft predictive collision penalty mechanism to reduce collision rates by penalizing obstacle-approaching actions. 
To ensure that the trained reinforcement learning network can directly transfer to real-world scenarios, we have established a Real2Sim2Real simulator. 
In the Real2Sim step, we use 3D Gaussian Splatting (3DGS) to transform real-world scenes into digital scenes.
In the Sim2Real step, we deploy the end-to-end model onto the vehicle to bridge the Sim2Real gap.
Trained in the 3DGS simulator and deployed on physical vehicles, 
REAP successfully parks in various types of parking spaces, especially demonstrating the feasibility of end-to-end RL parking in extremely narrow mechanical slots.


\end{abstract}

\section{Introduction}
\label{sec:intro}

\IEEEPARstart{T}{he} end-to-end paradigm in autonomous driving has become a hot research topic due to its advantages of minimal cumulative error and reduced information loss between modules. 
Most end-to-end algorithms in autonomous driving adopt imitation learning methods, such as \cite{prakash2021multi, hu2023planning, chen2024vadv2, li2024hydra, liao2025diffusiondrive, li2024parkinge2e}, aiming to acquire driving capabilities by learning from human drivers. 
However, imitation learning algorithms require vast amounts of high-quality training data, resulting in significant data production costs and resource consumption.
Additionally, diverse expert trajectories can also lead to an averaging behavior problem in imitation learning results, making it difficult to effectively complete different tasks.
Multi-stage rule-based planning methods \cite{noto2000method, Kurzer1057261, tang2021geometric, he2022dynamic, noreen2016optimal, bohlin2000path, sussmann1991shortest} do not require expert data, but these approaches often rely on pre-built maps or real-time perception results.
This often leads to planning failures in complex scenarios when there is no prior map or when real-time perception is inaccurate.


\begin{figure*}[ht]
	\centering
        \includegraphics[width=0.9\linewidth]{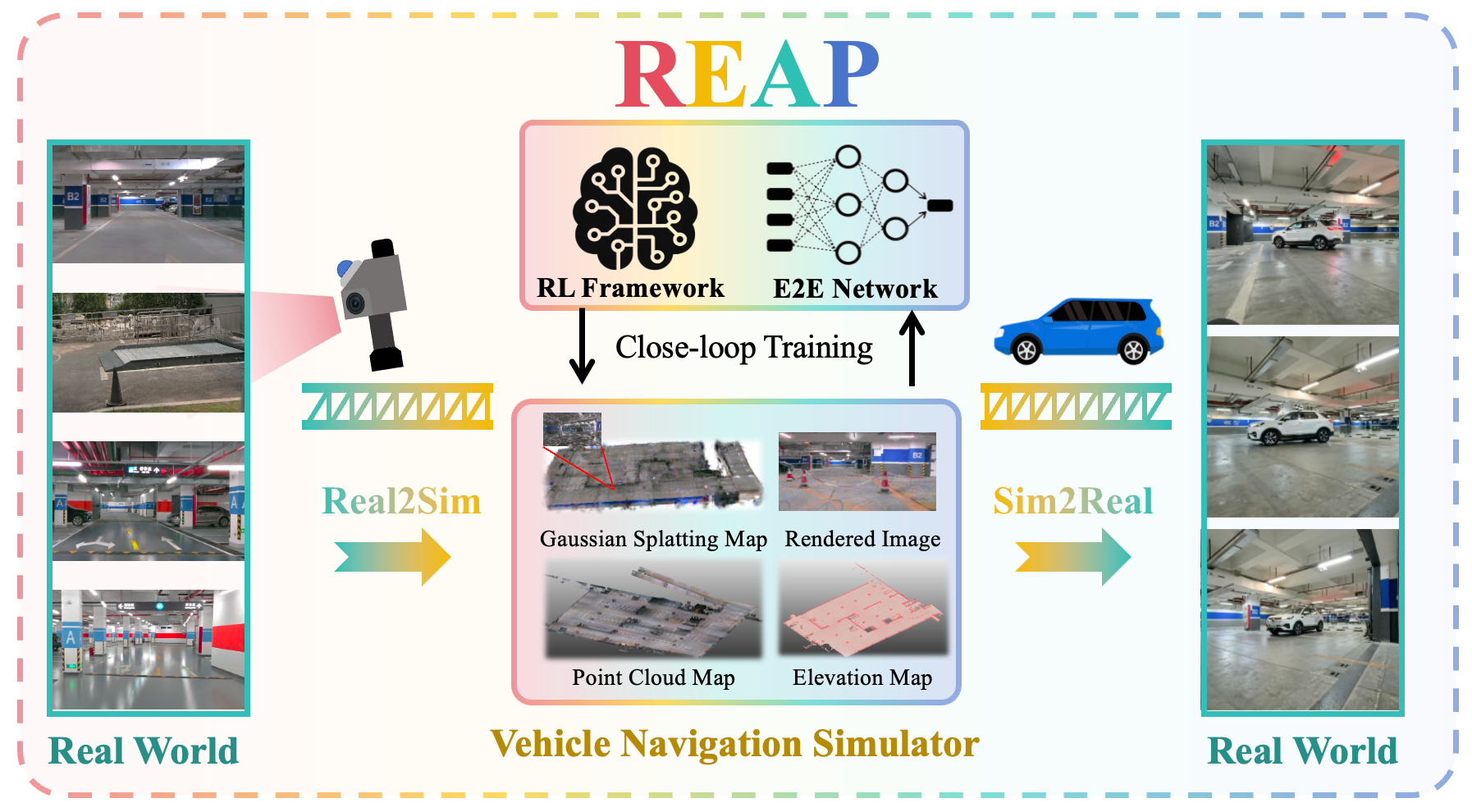}
	\caption{\textbf{The structure of the overall workflow.} The algorithm adopts the Real2Sim2Real paradigm. \textbf{Real2Sim}: Construct simulation scenarios using data collected via handheld devices. \textbf{Closed-loop training}: Train the network model in the simulator using an end-to-end reinforcement learning approach. \textbf{Sim2Real}: Deploy the model trained in the simulator to a real vehicle, enabling real-world validation in standard garages and mechanical parking spaces.}
	\label{fig:abs}
\end {figure*}

Deep reinforcement learning (DRL) \cite{sutton1998reinforcement}, as an interactive learning method, has achieved remarkable success in scenarios such as games and gambling.
In the field of robotics, reinforcement learning is often trained in simulators \cite{dosovitskiy2017carla, makoviychuk2021isaac, todorov2012mujoco, juliani2018unity}.
However, a Sim2Real gap remains, leading to significant degradation in model performance when models trained in simulators are transferred to real-world environments.
Recently, some methods \cite{jia2025discoverse, zhu2025vr} have adopted 3D Gaussian Splatting (3DGS) combined with physics engines \cite{makoviychuk2021isaac, todorov2012mujoco} to achieve visual fidelity. However, these simulators mainly focus on manipulator arms and legged robots and still lack effective Real2Sim2Real simulation platforms for wheeled robots.
There are also methods that bypass simulators and directly build reinforcement learning algorithms in real-world scenarios \cite{luo2024precise}.
However, these approaches often face challenges such as complex setup requirements and the need for precise environmental resets, which can limit their widespread application.

Addressing current challenges faced by end-to-end autonomous driving algorithms and reinforcement learning, we propose a vision-based Reinforcement learning End-to-end Autonomous Parking (REAP) algorithm that predicts the vehicle's actions through surround-view camera images.
The overall workflow of our algorithm is shown in Fig. \ref{fig:abs}.

By combining the end-to-end network architecture with the proposed training paradigm, our algorithm achieves precise parking even in challenging parking spaces.


Our main contributions are summarized as follows:
\begin{itemize}
    \item We present \textbf{REAP}, a vision-based \textbf{R}einforcement learning-based \textbf{E}nd-to-end \textbf{A}utonomous \textbf{P}arking algorithm, significantly enhancing parking capabilities in extreme scenarios. 
    The algorithm clones the rule-based planning capabilities into the model's capabilities through behavior cloning and adopts asymmetric reinforcement learning by feeding privileged ground-truth information into the critic network, effectively improving the training efficiency of reinforcement learning.

    \item We design a closed-loop vehicle navigation simulator based on 3DGS, effectively narrowing the Sim2Real gap by integrating various transfer strategies. The simulator adopts a Real2Sim2Real pipeline, achieving the direct applicability of simulation-trained parking models to real-world scenarios.
    
    \item To enhance safety in autonomous parking, we propose a soft predictive collision penalty mechanism based on distance fields and predictive collision assessment, effectively reducing collision rates.

    \item We successfully deploy our algorithm on a real-world vehicle platform, validating its parking capability and precision in extreme scenarios such as mechanical parking slots. To the best of our knowledge, this is among the first end-to-end reinforcement-learning approaches to achieve parking in extremely narrow mechanical slots.
\end{itemize}
    

\section{Related Works}
\label{sec:rela}


\subsection{End-to-end Autonomous Driving}
End-to-end autonomous driving replaces the modular perception–prediction–planning pipeline with a unified learning system. 
Compared to multi-stage approaches, end-to-end methods demonstrate advantages in both generalization and scalability.
Current end-to-end algorithms are primarily focused on driving tasks, with imitation learning being the dominant paradigm.
TransFuser \cite{prakash2021multi} integrates image and LiDAR representations using attention to handle adversarial scenarios.
UniAD \cite{hu2023planning} jointly optimizes detection, tracking, motion prediction, occupancy mapping, and planning under a unified network to reduce error propagation.
VADv2 \cite{chen2024vadv2} models a probabilistic distribution over candidate trajectories, tokenizing the planning space into a large planning vocabulary.
Hydra-MDP \cite{li2024hydra} introduces a teacher-student knowledge distillation framework where the student model learns from both human driving demonstrations and rule-based planners, enabling multi-target and multimodal planning in a fully end-to-end manner.
Recently, some algorithms have integrated generative models into end-to-end frameworks, alleviating the averaging behavior problem in trajectory planning and decision-making, thereby enhancing algorithm performance.
DiffusionDrive \cite{liao2025diffusiondrive} introduces a novel end-to-end autonomous driving policy based on a truncated diffusion model, which starts its denoising process from an anchored Gaussian distribution to generate diverse multimode trajectories more efficiently.

Although end-to-end algorithms have shown promising performance in structured road driving scenarios, there is still limited research on their application to autonomous parking. Compared to continuous driving, parking scenarios suffer from severe data imbalance. If the predictions at critical gear-shift points are inaccurate, it often leads to ultimate parking failure.
ParkingE2E \cite{li2024parkinge2e} tackles the task of autonomous parking using a camera-based end-to-end neural network. 
However, the consistency of expert parking trajectories is particularly difficult to guarantee, making end-to-end parking more prone to the averaging behavior and mode collapse problems compared to regular driving tasks. 
Furthermore, the scarcity of high-quality parking datasets makes it challenging to ensure the robust generalizability of these end-to-end parking algorithms.

\begin{figure}[h]
	\centering
        \includegraphics[width=1.0\linewidth]{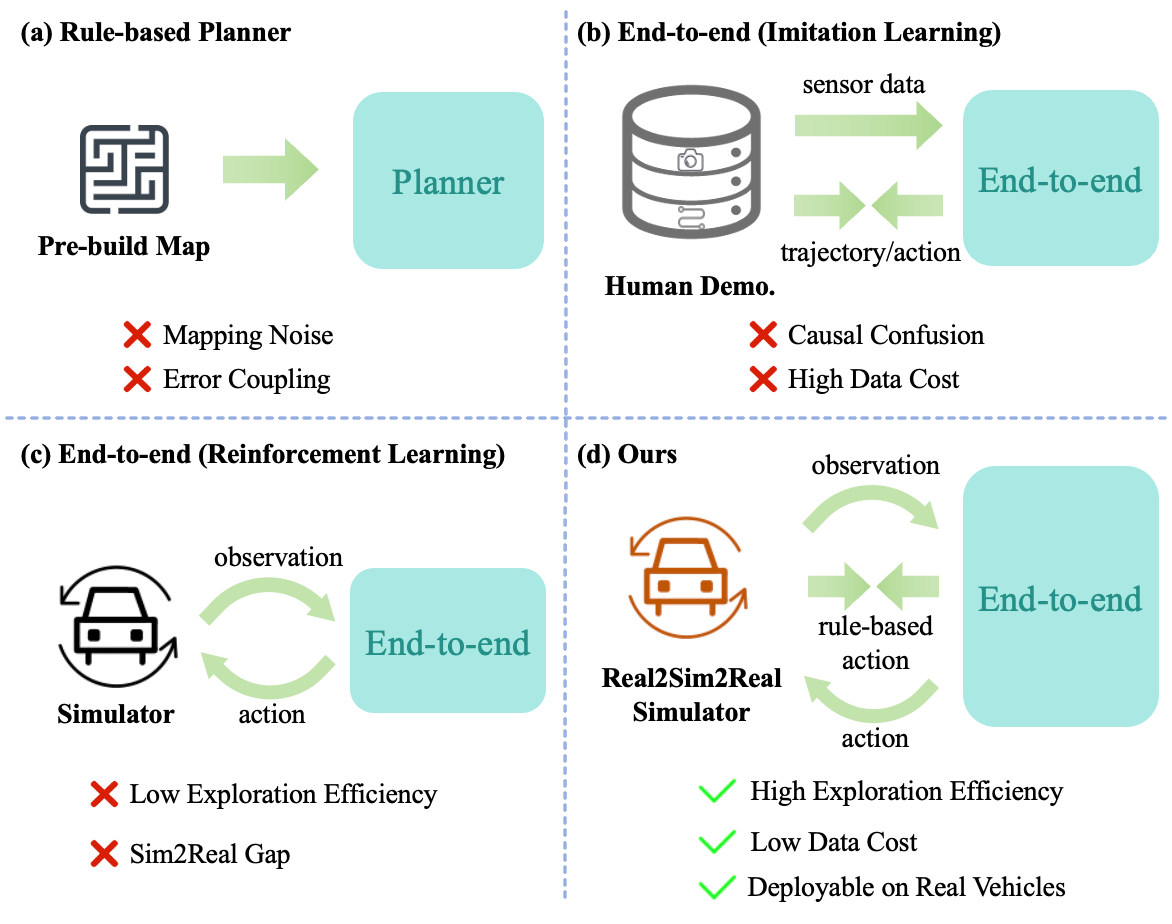}
	\caption{\textbf{The comparison of different methods.} (a) Rule-based planner. (b) Imitation learning end-to-end method. (c) Reinforcement learning end-to-end method. (d) Our proposed method.}
	\label{fig:compare}
\end {figure}

A high-level comparison of the main methodological paradigms discussed above is shown in Fig.~\ref{fig:compare}.

\subsection{Reinforcement Learning for Autonomous Parking}
Compared to imitation learning, which relies on expert demonstrations, reinforcement learning offers better scalability and adaptability for autonomous parking.
Common reinforcement learning training methods include DDPG \cite{lillicrap2015continuous}, PPO \cite{schulman2017proximal}, SAC \cite{haarnoja2018soft}, and TD3 \cite{dankwa2019twin}.
Recently, RL has demonstrated significant potential in tackling complex autonomous driving challenges, ranging from multi-objective strategic planning \cite{tong2025multi} to end-to-end maneuver control in demanding scenarios like consecutive sharp turns \cite{li2025investigation}.

For camera-based autonomous parking, learning policies directly from high-dimensional pixel inputs further increases the difficulty and instability of reinforcement learning.
Although vision-based end-to-end methods \cite{zhao2023learning} have been extensively studied in autonomous driving and robotics, most existing algorithms still rely on imitation learning because of the high training cost of reinforcement learning and the transfer challenges caused by the visual gap.
A common strategy in this setting is offline reinforcement learning \cite{fang2022offline, li2024boosting}, which improves training efficiency and has therefore attracted widespread research interest.

In the parking domain, reinforcement-learning-based methods have also been explored.
For example, HOPE \cite{jiang2025hopereinforcementlearningbasedhybrid} studies a hybrid policy path planner for diverse parking scenarios, RL-OGM-Parking \cite{wang2025rlogmparkinglidarogmbasedhybrid} investigates a LiDAR OGM-based hybrid reinforcement learning planner with real-world validation, while SEG-Parking \cite{li2024segparking} explores safe and generalizable autonomous parking via end-to-end offline reinforcement learning. 

\subsection{3D Gaussian Splatting Simulator}
3D Gaussian Splatting (3DGS) \cite{kerbl20233d} introduces an explicit, anisotropic Gaussian scene representation that enables real-time, high-resolution novel-view synthesis and fast optimization, providing photorealistic rendering.
Some early methods \cite{wu2023mars, wu2025rlgsbridge3dgaussiansplatting, yang2023unisim} utilized NeRF \cite{mildenhall2021nerf} to construct sensor simulators, demonstrating notable advantages in novel-view synthesis.
Building upon this, recent works have developed 3DGS-based simulators for robotic manipulation \cite{jia2025discoverse} and legged locomotion \cite{zhu2025vr}. By integrating 3DGS with physics engines, these platforms achieve coupled high-fidelity visual and physical simulation, significantly accelerating Sim2Real deployment. 
However, high-fidelity visual-physical simulators tailored to autonomous vehicles are still lacking.
While established environments like CARLA \cite{dosovitskiy2017carla} provide robust vehicle dynamics and broad scenarios, they rely on traditional CAD assets that suffer from a noticeable visual domain gap, hindering direct Sim2Real transfer without complex domain adaptation.



\begin{figure*}[t]
	\centering
	\includegraphics[width=\textwidth]{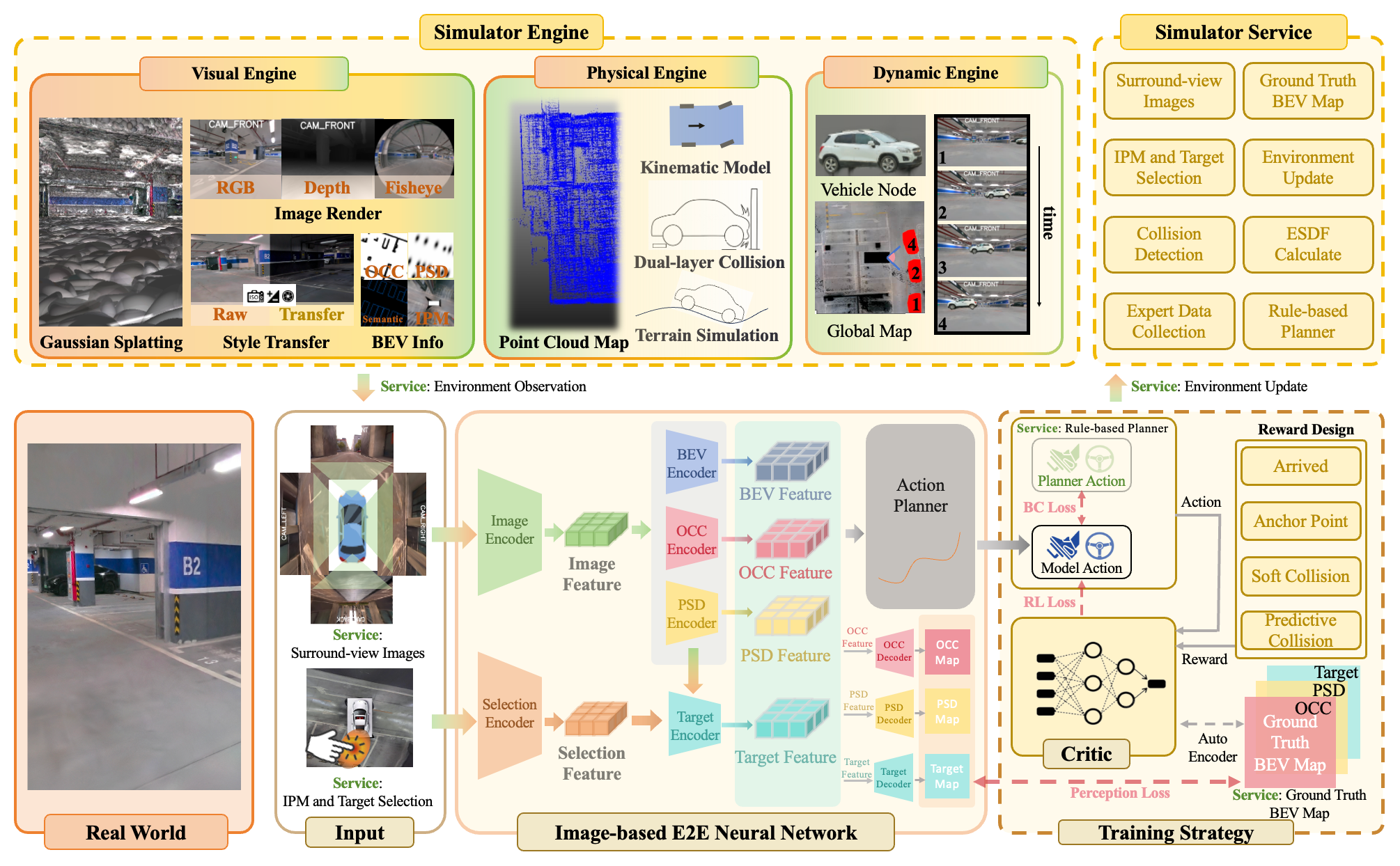}
	\caption{\textbf{Overview of the proposed end-to-end autonomous parking framework}. The system utilizes surround-view images and user-selected target heatmaps as inputs, which are encoded into latent spatial features. These features are fused and fed into the actor network for continuous action prediction. During training, auxiliary BEV perception tasks are co-optimized to enhance feature representation, and ground-truth BEV maps serve as privileged information for the critic network to guide efficient policy learning. To bridge the Sim2Real gap, our Real2Sim2Real pipeline leverages a 3DGS-based simulator enhanced with domain randomization, enabling the policy trained entirely in simulation to directly generalize to real-world physical deployment.
		}
	\label{fig:framework}
\end{figure*}

\section{Task Definition}
Our task is to accomplish autonomous parking for intelligent vehicles. The approach employs a vision-based solution, utilizing surround-view camera images as input and vehicle actions as output.

\textbf{Inference Process}: 
The surround-view cameras capture images that are transformed into Inverse Perspective Mapping (IPM) images. 
The user interacts with an IPM image to select the target parking space, thereby generating a target heatmap in the Bird's-Eye View (BEV) space. 
Both the surround-view images and the target heatmap serve as input to the parking network. Through feature encoding and action decoding, the network predicts parking actions.

\textbf{Training Process}: 
We use a total of $n$ training scenes, and scene $i$ contains $l_i$ parking slots. 
During each training episode, we randomly select a target parking space from one of the scenes and place the vehicle at a collision-free location within a certain range of the target space, serving as the vehicle's starting position.
The surround-view images rendered from the vehicle's starting position, together with a perturbed target heatmap of the selected parking space, are fed into the parking network to predict vehicle actions.

\section{Methodology}
\label{sec:methodology}
\subsection{End-to-end Network with BEV Auxiliary Tasks}
\label{auxiliary_tasks}
The framework of our proposed method REAP is shown in Fig. \ref{fig:framework}. 
In our pipeline, we adopt a unified BEV feature representation.
The set of images from $N$ surround-view cameras $C = \{\textbf{c}_i| i \in  \mathbb{N}, 1 \leq i \leq N\}$ are transformed into the BEV view \cite{philion2020lift}, while the target input is also represented using BEV target heatmap $T$, ensuring a unified spatial representation. The BEV features of images and selected target heatmap features are denoted as $f_\mathrm{image}$ and $f_\mathrm{selection}$, respectively.
We integrate auxiliary tasks to extract important BEV features. Occupancy task (OCC task) provides effective geometric occupancy information, helping the planning module avoid obstacles.
Parking slot detection task (PSD task) predicts the available parking spots in the scene, delivering a stable perception of parking spaces.
Target prediction task (Target task) combines the results of the PSD task with the input target heatmap to produce a stable prediction of the target parking slot.
If rectangular boxes are used to represent parking slot, multiple adjacent parking spaces may merge into one, causing different parking areas to lose their distinctiveness. 
We adopt a 2D Gaussian distribution to model the parkable slot, where the function values decrease significantly near the edges of a parking space, thereby ensuring good separability between different spaces. 
Additionally, to distinguish the directions into which a vehicle can park, we employ an asymmetric gating function to differentiate between accessible and inaccessible parking directions.

We first use a BEV mapping module to extract image features with both spatial and semantic information, formulated as $f_\mathrm{BEV} = \mathcal{F}_\mathrm{BEV}(f_\mathrm{image})$.
The occupancy task encodes $f_\mathrm{BEV}$ into occupancy features $f_\mathrm{OCC} = \mathcal{F}_\mathrm{OCC}^\mathrm{enc}(f_\mathrm{BEV})$, which are then fed into a decoder to predict the occupancy map $M_\mathrm{OCC}^\mathrm{pred} = \mathcal{F}_\mathrm{OCC}^\mathrm{dec}(f_\mathrm{OCC})$.
Similarly, the parking slot detection features and the parking slot detection map are denoted as $f_\mathrm{PSD}$ and $M_\mathrm{PSD}^\mathrm{pred}$, respectively.
The target feature is computed as $f_\mathrm{target} = \mathcal{F}_\mathrm{target}^\mathrm{enc}(f_\mathrm{PSD}, f_\mathrm{selection})$ and fed into a decoder to predict the target map $M_\mathrm{target}^\mathrm{pred}$.

The loss function $\mathcal{L}_\text{aux}$ of the perception module is formulated as follows:
\begin{equation}
    \mathcal{L}_\text{aux} = \mathcal{L}_\text{OCC} + \mathcal{L}_\text{PSD} + \mathcal{L}_\text{target},
\end{equation}
where $\mathcal{L}_\text{OCC}, \mathcal{L}_\text{PSD}$, and $\mathcal{L}_\text{target}$ denote the losses between the ground-truth and predicted maps for occupancy, parking-slot detection, and target prediction, respectively.
These losses for auxiliary tasks are computed using the MSE (Mean Squared Error) loss. 

We then fuse these features in the action-decoding module, which outputs predicted vehicle actions (speed and steering). These actions are either used to update the simulated vehicle state or executed by the real vehicle actuators.
\begin{equation}
    a=\mathcal{F}_\mathrm{planner}(f_\mathrm{OCC}, f_\mathrm{PSD}, f_\mathrm{target}, f_\mathrm{BEV}),
\end{equation}
where $\mathcal{F}_\mathrm{planner}$ represents the neural network of the planner and $a$ represents the predicted action. In stochastic reinforcement learning, actions are sampled from a Gaussian distribution.

The loss function of the end-to-end network can be represented as:
\begin{equation}
    \mathcal{L} = \mathcal{L}_\text{action} + \beta \mathcal{L}_\text{aux},
\end{equation}
where $\beta$ represents the weighting coefficient of the auxiliary task loss.
The action loss $\mathcal{L}_\text{action}$ can be denoted as:
\begin{equation}
\mathcal{L}_\text{action} =
\begin{cases}
\mathcal{L}_\text{action}^\text{BC}, & 
\text{if the rule-based action is available}\\
& \text{and is used with probability } p, \\[4pt]
\mathcal{L}_\text{action}^\text{RL}, & \text{otherwise.}
\end{cases}
\end{equation}
where $\mathcal{L}_\text{action}^\text{BC}$ is defined as the MSE loss between the predicted actions and the rule-based (desired) actions, and $\mathcal{L}_\text{action}^\text{RL}$ is determined by the specific reinforcement learning algorithm.
If the action originates from the rule-based planner and is used with probability $p$, the action loss $\mathcal{L}_\text{action}$ is expressed as $\mathcal{L}_\text{action}^\text{BC}$. Otherwise, the action loss $\mathcal{L}_\text{action}$ is expressed as $\mathcal{L}_\text{action}^\text{RL}$.
The definitions of $\mathcal{L}_\text{action}^\text{BC}$ and $\mathcal{L}_\text{action}^\text{RL}$ are given in Section \ref{strategy}. For the selection criterion of the probability $p$, please refer to Section \ref{details}.


\subsection{Asymmetric Hybrid Reinforcement Learning Strategy}
\label{strategy}
Reinforcement learning theoretically finds optimal policies, but vision-based parking suffers from inefficient exploration and complex reward design. 
Although offline reinforcement learning or standard distillation improves efficiency, they require massive static pre-collected demonstration datasets and often suffer from distribution shift during online execution. 
To address these limitations, we propose a dynamic hybrid strategy that combines online reinforcement learning with behavior cloning. 
Unlike offline approaches, our method dynamically queries a rule-based planner during exploration to generate in-distribution expert supervision. 
This provides immediate, state-matched guidance, thereby avoiding covariate shift and significantly accelerating the online training process without requiring pre-collected datasets.
In addition, the asymmetric design lies in assigning different observation modalities to the actor and critic: the actor learns directly from surround-view images, while the critic is trained with privileged artifact-free BEV states available only in simulation. This design improves value-estimation stability, increases training efficiency, and reduces critic-side computational cost because the BEV input is more compact than raw surround-view images and can be processed with a lighter critic backbone.

The training strategy follows an online learning paradigm. The action sources include two types: model-predicted actions and rule-based actions. When rule-based actions are not available, model-predicted actions are used to collect data, and the collected data are used for reinforcement learning. When rule-based actions are available, behavior cloning is applied with probability $p$.

In the training stage, reinforcement learning employs the soft actor-critic (SAC) algorithm, a maximum-entropy reinforcement learning method that maximizes both the action value and the policy entropy. 
The actor network is responsible for predicting vehicle actions in an end-to-end manner, while the critic network is used to evaluate the policy.
The actor network comprises a feature encoder and an action-planning module, as described in Section \ref{auxiliary_tasks}.
The actor loss function of SAC can be expressed as:
\begin{equation}
    \mathcal{L}_\text{action}^\text{RL} = -\mathbb{E}_{s_t \sim D} [-\alpha \log \pi_{\theta}(a_t | s_t) + Q(s_t, a_t)],
\end{equation}
where $\theta$ denotes the network parameters, $D$ denotes the data distribution, and $Q(s_t, a_t)$ represents the action-value function, which predicts the expected cumulative reward of taking action $a_t$ in state $s_t$. The entropy term, $-\log\pi_{\theta}(a_t | s_t)$, promotes policy diversity and encourages exploration, and $\alpha$ denotes the temperature coefficient.

We employ Reed-Shepp curves as the rule-based trajectory planner.
During exploration, if a feasible path is found, we use the planner's actions to collect data.
However, these actions are not predicted by the neural network. During the reinforcement learning training process, the network's output from the actor network is indirectly influenced by increasing the value of these actions. This often makes it inefficient for the network to learn such actions.
We address this problem by distilling the action policy of the rule-based planner into the actor network through behavior cloning. The loss function of the behavior clone is expressed as:
\begin{equation}
\mathcal{L}^{\text{BC}}_{\text{action}} = \mathbb{E}_{s_t \sim D} \left[ \left\| \pi_\theta(s_t) - a_t^{\text{ref}} \right\|^2 \right],
\end{equation}
where $a_t^\mathrm{ref}$ is the reference action.

The critic network consists of feature encoding and policy evaluation components. 
Vision-based policy evaluation inherently suffers from the Partially Observable Markov Decision Process (POMDP) problem, leading to high-variance value estimates. 
To mitigate this and increase training stability, we adopt an asymmetric reinforcement learning architecture. 
Instead of surround-view camera images, privileged artifact-free BEV maps (occupancy map, parking slot detection map, target map) from the simulator are fed directly into the critic network. 
These BEV maps are processed through an autoencoder, providing the critic with a stable, fully-observable state representation to accurately guide the visual actor. 
These latent encoder features, along with the actor-predicted action feature, are fed into the critic module.
The loss function of the critic network consists of the feature auto encoder loss and the value loss. 
In the critic network, actions from the rule-based planner and network prediction are not distinguished.

Unlike the standard SAC algorithm, which employs a single replay buffer, the proposed method utilizes two replay buffers. The first is the primary replay buffer, into which all experience data are stored. The second is a successful-experience replay buffer, which specifically stores the experiences from episodes that achieve collision-free success. During training, samples are drawn from both buffers according to a predefined ratio. This design is motivated by the observation that, in extremely challenging parking scenarios, successful experiences are difficult to obtain. By introducing the successful-experience replay buffer, the utilization efficiency of such rare but valuable experiences can be significantly improved.

\subsection{Reward Design}
\label{collision_penalty}
In the design of parking algorithms, critical metrics include whether the vehicle can successfully park in the target slot and whether the process is collision-free and time-efficient.
Based on these metrics, we design the reward $R_\text{total}$, which consists of two parts: sparse rewards and dense rewards. The $R_{\text{total}}$ can be represented as:
\begin{equation}
R_{\text{total}} = R_{\text{sparse}} + R_{\text{dense}},
\end{equation}
where sparse rewards $R_\text{sparse}$ are primarily used to evaluate the overall performance of the current episode:
\begin{equation}
R_{\text{sparse}} = R_{\text{succ}} + R_{\text{PC}} + R_{\text{anchor}} + R_{\text{timeout}} + R_{\text{bound}},
\end{equation}
which includes successful parking reward $R_\text{succ}$, predictive collision reward $R_\text{PC}$, parking anchor reward $R_\text{anchor}$, timeout penalty $R_\text{timeout}$, and out-of-bounds penalty $R_\text{bound}$. 

We also adopt dense rewards $R_\text{dense}$:
\begin{equation}
R_{\text{dense}} = R_{\text{IoU}} + R_{\text{SC}},
\end{equation}
which include Intersection-over-Union (IoU) reward $R_\text{IoU}$ and soft collision reward $R_\text{SC}$.
Together, the predictive collision reward $R_\text{PC}$ and the soft collision reward $R_\text{SC}$ constitute the proposed soft predictive collision penalty mechanism. The two terms play complementary roles: $R_\text{PC}$ evaluates the predicted collision extent along the action trajectory and penalizes unsafe actions before a hard collision is fully completed, while $R_\text{SC}$ penalizes overly small clearance to nearby obstacles and explicitly encourages a safer parking margin even when no terminal collision occurs.

\textbf{Successful parking reward}: 
During training, if the distance and angle between the vehicle and the target parking space are within certain threshold ranges, the parking is considered successful. The agent then receives a high success reward $R_\mathrm{succ}$, and the current episode is terminated.

\begin{figure}[t]
	\centering
    \includegraphics[width=0.9\linewidth]{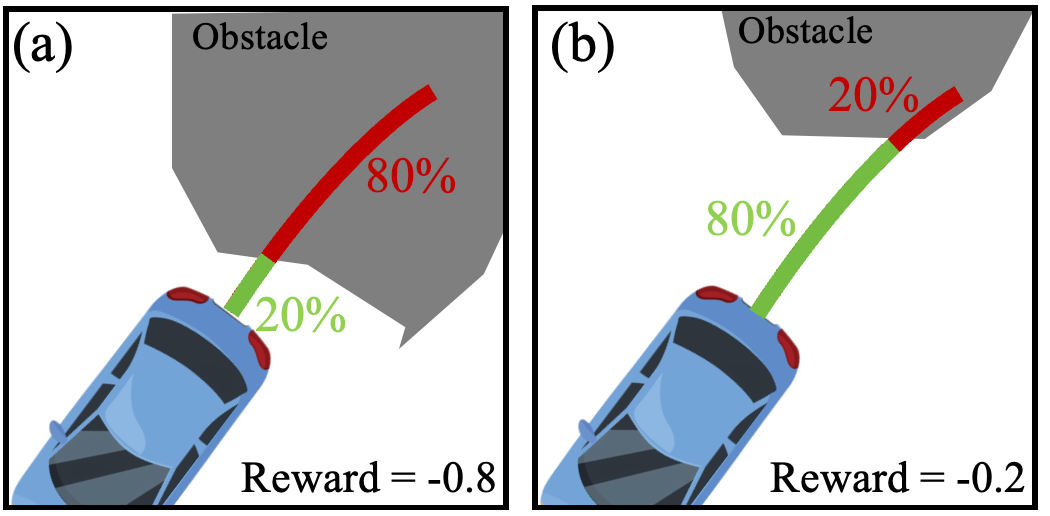}
	\caption{\textbf{Different collision scenarios.} Collision status of the vehicle and its predicted actions. Red indicates collision areas, and green indicates collision-free areas. (a) 80\% of the trajectories result in collision; (b) 20\% of the trajectories result in collision.}
	\label{fig:predictive_collision_reward}
\end {figure}

\textbf{Predictive collision reward}: 
Standard reinforcement learning methods typically assign a fixed negative reward for any collision. However, this uniform penalty fails to reflect the actual severity of different collisions, limiting the policy's ability to learn safe behaviors.
To resolve this, we propose a predictive collision reward $R_\text{PC}$. 
$R_\text{PC}$ provides a continuous reward by calculating the ratio of the collision length to the total length of the action trajectory. 
This continuous scaling enables the policy to distinguish more severe predicted collisions from milder ones and guides it away from catastrophic actions before they fully occur. 
Different collision scenarios are shown in Fig. \ref{fig:predictive_collision_reward}.

\textbf{Parking anchor reward}:
Parking is a long-horizon exploration task. If the reward is solely based on the discounted success reward propagated back to the initial state, the value estimation for states and actions far from the goal is often poorly learned. 
On the other hand, introducing dense rewards based on the vehicle's distance and orientation to the target slot can lead to undesirable behaviors due to the vehicle's kinematic constraints—specifically, the agent may attempt to maximize immediate rewards by frequently changing between forward and reverse gears, resulting in jerky, unnatural motion that severely degrades the parking experience.
In this paper, we designate all starting positions for which a valid rule-based planner generates a feasible trajectory as parking anchor points. When the vehicle reaches any of these anchor points during an episode, it receives an anchor point reward $R_\mathrm{anchor}$. Importantly, within each parking episode, the parking anchor reward is granted at most once.
It should be noted that if rule-based planner actions are not obtained during the parking process but the vehicle successfully parks on the slot, the parking anchor reward cannot be obtained in that episode. To encourage such successful parking behavior, this reward is compensated to the final action.

\textbf{Soft collision reward}: 
While the predictive penalty evaluates potential collisions, RL policies may still learn to drive excessively close to obstacles. 
To encourage larger safety margins, we introduce a soft collision reward. 
This mechanism acts as a safety buffer by using 2D occupancy grid maps to generate a truncated signed distance field (TSDF): 
\begin{equation}
    \label{tsdf}
    \mathcal{T}(\textbf{p}) = \max[0, (\frac{\min[D(\textbf{p}), d_0]}{d_0})^\tau],
\end{equation}
where $\mathcal{T}$ represents the function of TSDF, $\textbf{p}$ is the pixel point on the occupancy grid maps, $D(\textbf{p})$ denotes the Euclidean distance between the point and the closest obstacle, $d_0$ is the threshold of truncated Euclidean distance, and $\tau$ is the non-linear factor to adjust the gradient of TSDF. 
The specific implementation settings of $d_0$ and $\tau$ used in our experiments are provided in Section~\ref{details}.
The vehicle state before taking the action is denoted as $\textbf{s}_0$.
After taking the action, the process from the initial state to the final state can be discretized into $n$ states, denoted as $\textbf{S}=\{\textbf{s}_0, \textbf{s}_1, ...,\textbf{s}_{n-1}\}$.
$B(\textbf{s})$ represents the set of points on the vehicle boundary defined in a specific vehicle state $\textbf{s}$.
The soft collision reward $R_\text{SC}$ can be represented as:
\begin{equation}
\label{sc_reward}
    R_\text{SC}=\underset{\textbf{s}\in \textbf{S}}{\min}\underset{\textbf{p} \in B(\textbf{s})}{\min}[\mathcal{T}(\textbf{p})] - \underset{\textbf{p} \in B(\textbf{s}_0)}{\min}[\mathcal{T}(\textbf{p})].
\end{equation}
This formula indicates that the soft collision reward is determined jointly by the initial state and the state corresponding to the minimum distance to the obstacles.
The soft collision reward encourages the vehicle to maintain a safe distance $d_0$ from obstacles, thereby significantly reducing collision probabilities and avoiding overly aggressive maneuvers with insufficient clearance.


\begin{figure*}[h]
	\centering
        \includegraphics[width=\linewidth]{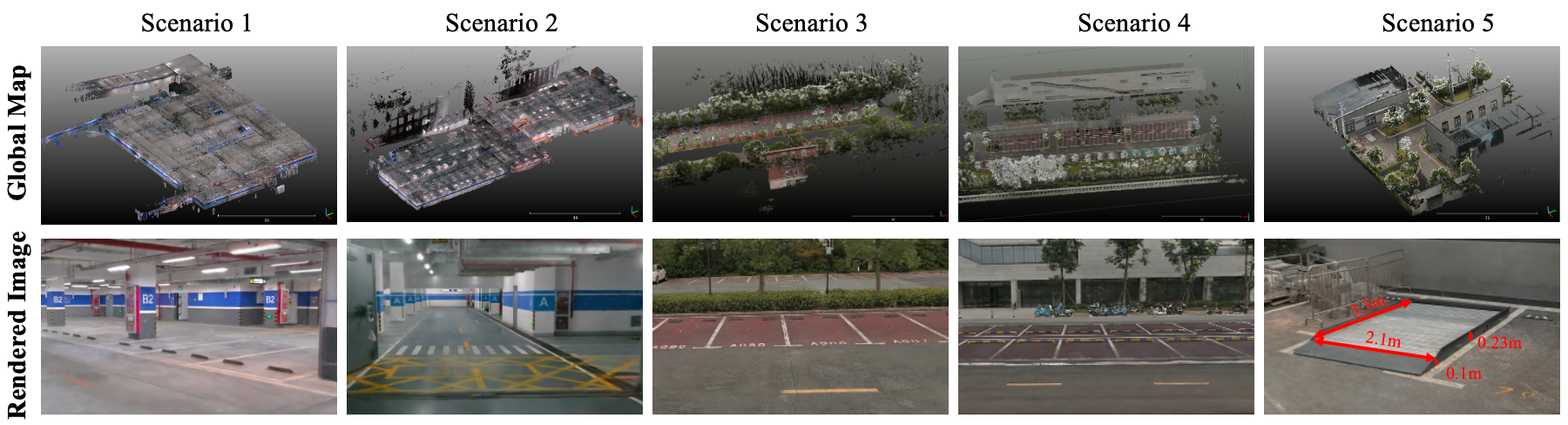}
	\caption{\textbf{Description of the scenarios used in training and validation}. Scenario 1 (approximately 60 parking spaces) and Scenario 2 (approximately 100 parking spaces) are both underground parking slots. Scenario 3 (approximately 60 parking spaces) and Scenario 4 (approximately 60 parking spaces) are ground parking slots. Scenario 5 is a mechanical parking space, with a width of only 2.1 m and side panels on both sides. When a vehicle enters the mechanical parking space, there is a clearance of only about 10 cm between the vehicle and the side panels, making parking in the mechanical scenario a highly challenging task.}
	\label{fig:scene}
\end{figure*}

\subsection{GS-based Vehicle Navigation Simulator}
We constructed a Real2Sim2Real vehicle navigation simulator based on 3D Gaussian Splatting (3DGS) and point cloud maps to adapt our camera-based end-to-end parking algorithm for RL training and transfer to the real world.
We collect sensor data from the real environment to construct simulation scenarios, achieving the Real2Sim conversion.
Models trained in the simulator can then be deployed on real vehicles for validation, enabling Sim2Real transfer.

\textbf{Real2Sim}: 
In the Real2Sim stage, we aim to accurately capture the texture and geometric information of real-world scenes. 
We utilize a handheld LiDAR scanner to efficiently collect environmental data, avoiding manual Computer-Aided Design (CAD) modeling overhead and ensuring strong scalability. 
By using the generated precise LiDAR point clouds as initialization, we significantly accelerate 3DGS training and alleviate the geometric inaccuracies of the standard 3DGS method. 
This achieves high-fidelity, scale-accurate scene reconstruction. 

\textbf{Sim2Real}:
During the transfer process from the simulator to the real-world environment, there are typically two types of gaps: perception gap and physical gap.
\subsubsection{GS-based camera rendering}
For vision-based end-to-end training, the perception gap primarily arises from insufficient visual rendering fidelity in the simulator.
To address this, we employ a rendering approach based on Gaussian splatting. Specifically, we apply random perturbations to image styles by adjusting exposure and performing domain randomization, while also adding stochastic noise to camera intrinsics and extrinsics.



\subsubsection{Kinematics-based vehicle motion}
Many current end-to-end trajectory prediction models face the challenge of ensuring that generated trajectories adhere to kinematic constraints, making the predicted trajectories potentially infeasible in real-world applications. 
Our algorithm predicts vehicle actions $a = (v, \omega)$, where $v$ defines the velocity of the ego vehicle and $\omega$ indicates the steering angle of the front wheels. We convert them into executable trajectories using a bicycle model:
\begin{equation}
\begin{aligned}
    &\psi(\Delta t) = \int_0^{\Delta t} \frac{v}{L} \tan(\omega) dt, \\
    &x(\Delta t) = \int_0^{\Delta t} v \cos(\psi(t)) dt, \\ 
    &y(\Delta t) = \int_0^{\Delta t} v \sin(\psi(t)) dt, \\
\end{aligned}
\label{bycicle_model}
\end{equation}
where $x(\Delta t)$, $y(\Delta t)$ and $\psi(\Delta t)$ define the 2D position and orientation angle of the vehicle after $\Delta t$ seconds relative to the current ego vehicle coordinate system. $L$ denotes the wheelbase of the vehicle.

To enable the vehicle to adapt to ground slopes and perform precise collision detection, we utilize a 2.5D elevation map along with a dual-layer vehicle model. 
This approach allows for accurate simulation of the vehicle's movement over various terrains and obstacles, including: (1) Ascending and Descending Slopes: The 2.5D elevation map provides detailed height information, enabling the simulation of uphill and downhill driving.
(2) Speed Bumps and Low Obstacles: The model can accurately simulate the vehicle passing over speed bumps and low obstacles such as parking space speed poles.
(3) It also supports simulating entry and exit from mechanical parking spaces, which often have specific height and slope requirements.
(4) The system effectively handles collisions with standard obstacles like walls and pillars, ensuring realistic interaction within the environment.
The combination of the 2.5D elevation map and the dual-layer vehicle model enhances the precision of these simulations, making it possible to realistically replicate complex real-world scenarios in the simulator.




\section{Experiments}

\subsection{Scene Construction}
We use handheld devices to collect scene data.
Compared to vehicle-based collection, handheld devices provide a broader range of perspectives and more comprehensive observations, significantly enhancing the quality of scene construction and providing a high-quality environment for reinforcement learning exploration.
The sensors mounted on the handheld devices include an RGB camera, LiDAR, and IMU.
The geometric data from the point clouds serve as priors for the positions of the 3DGS Gaussian ellipsoids, allowing the construction of a 3DGS scene model with real-world scale.
To validate the algorithm's adaptability to different scenarios, we built multiple parking scenarios, including four parking-garage scenarios and one mechanical-parking scenario, providing approximately 300 available parking spaces in total. 
The specific scenes are shown in Fig. \ref{fig:scene}.
These parking garage scenarios exhibit substantial style variations, providing strong support for validating the algorithm’s generalization capability.

To quantitatively illustrate the parking difficulty in mechanical parking spaces, we provide the slot sizes, as shown in Fig. \ref{fig:scene}. The main challenges are as follows:
First, the parking slot is very narrow, with high side panels on both sides that cannot be encroached upon. The width of the mechanical parking slot is only 2.1 m, while the experimental vehicle’s width is approximately 1.9 m, leaving only about 0.1 m of clearance on each side.
Second, the mechanical parking slot has a slope of about 0.1 m. 
The rendered images differ noticeably between sloped and non-sloped cases. Ignoring this slope can lead to critical decision-making errors during parking, potentially causing parking failures.
Third, the length of the slot is shorter than the length of the vehicle. 
In collision detection within the simulator, if only a 2D occupancy map is used, the vehicle’s rear overhang cannot penetrate the barrier, leading to an inability to park in the slot. 
To address this, we use a 2.5D elevation map instead of a 2D occupancy map and construct a two-layer vehicle model  for collision detection. This allows the front and rear overhangs to penetrate low obstacles beneath them, which is crucial for parking in extremely confined mechanical parking spaces.

\subsection{Simulation Environment Setup}
We have developed a fully functional vehicle-navigation simulation environment.
Within the simulator, the vehicle is controlled through action commands (speed and steering), providing surround-view images, BEV maps (occupancy map, ramp map, parking slot map, target map, semantic map), vehicle pose, IPM maps, target parking spaces, and other relevant information.
The simulation environment supports both manual control and programmatic control modes.
To accelerate the simulation process, image rendering and BEV rendering are executed in separate threads.
Additionally, the simulation incorporates image noise, image style transformations, and random positional perturbations to prevent model overfitting to specific configurations.
Our simulator supports camera configurations with varying numbers of cameras, intrinsics, and extrinsics. 
To reduce the discrepancy between simulated and real-world camera rendering results, we configure the simulator’s visual sensors using the intrinsic and extrinsic parameters of the vehicle’s actual cameras.

\subsection{Implementation Details}
\label{details}
\textbf{Model Training}:
The model training process is divided into two stages: perception pretraining and end-to-end policy training.
In the first stage, vehicle positions are randomly generated in the simulator, with perturbations applied to the camera’s intrinsic and extrinsic parameters as well as to image styles.
The collected dataset is then used to pretrain the perception model.
In the second stage, the end-to-end model is trained with a hybrid strategy that combines reinforcement learning, rule-based guidance, and subsequent policy refinement.
For each training iteration, a parking slot within a scenario is randomly selected, and a random position and orientation within a certain range of that parking slot are sampled as the vehicle's starting pose.
The surround-view images rendered from this position, along with the perturbed target parking slot, are then fed into the network as input.
During policy training, we use whole-trajectory dropping, where the RS-trajectory drop probability $q$ follows a linear increase schedule as the number of training epochs increases.
Specifically, $q(t)=q_0+(q_{\max}-q_0)\min(t/T_{\mathrm{decay}},1)$, where $q_0=0$, $q_{\max}=1.0$, and $T_{\mathrm{decay}}=8{,}000$ epochs; equivalently, the probability of using rule-based actions is $p(t)=1-q(t)$.

\begin{figure}[t]
	\centering
	\includegraphics[width=\linewidth]{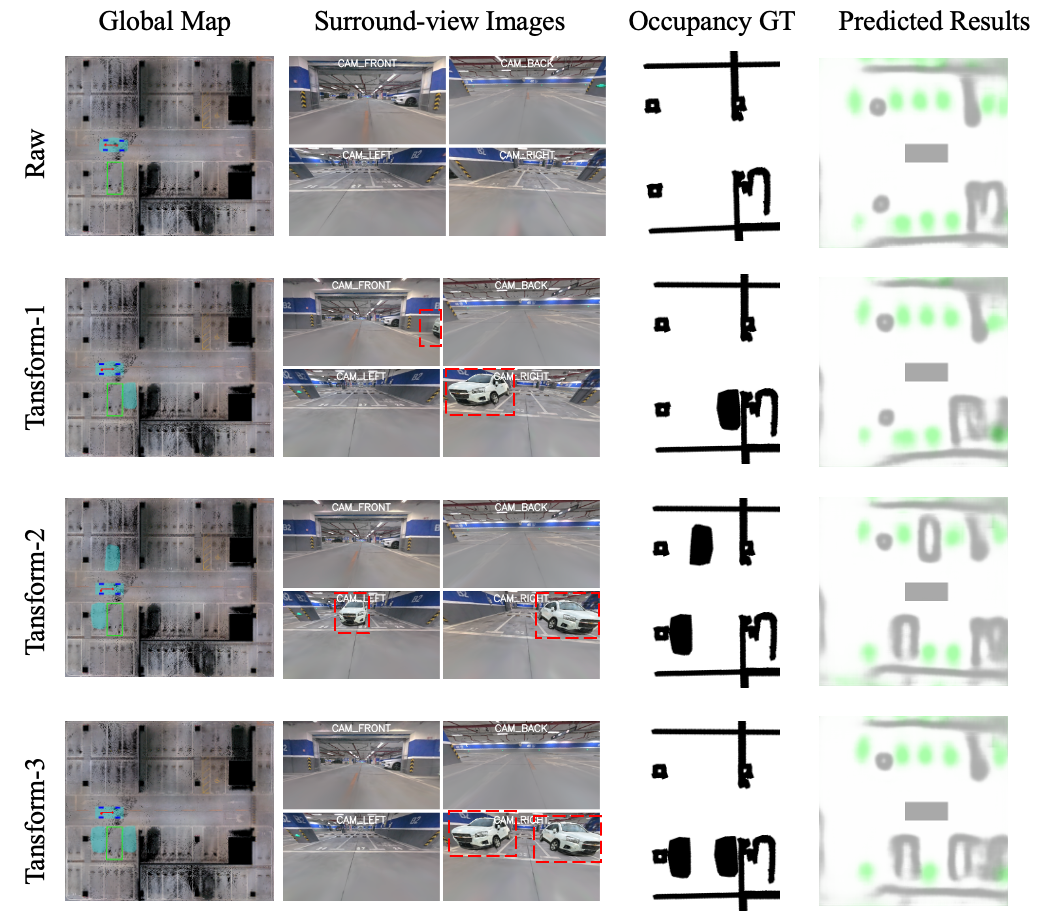}
	\caption{
		\textbf{Training scene randomization and visualization of auxiliary tasks prediction results.} The figure presents the global map, surround-view camera images, ground-truth occupancy map, and the composite visualization of occupancy and parking slot detection predicted by the model. The results demonstrate both the simulator’s capability of randomly adding vehicles and the model’s generalization ability in auxiliary tasks.
        The vehicle within the red dashed box in the surround-view images is a vehicle added by the simulator. In the BEV image, the vehicle's heading points to the right.
	}
	\label{fig:add_vehicle}
\end{figure}

Our constructed scenes contain a large number of empty parking spaces. 
However, in real parking scenarios, vehicles are inevitably present on both sides of the target parking space.
To prevent performance degradation under such conditions, we incorporated specific strategies during network training.
Specifically, we employed vehicle models built from the digital assets of the 3DRealCar \cite{du20243drealcar} dataset, with visual rendering based on Gaussian Splatting and geometric structure represented by point clouds.
During each training episode, 0 to 3 additional vehicles were randomly placed in the available parking spaces adjacent to the target space.
This design effectively simulates various real-world conditions.
Fig. \ref{fig:add_vehicle} shows the visualization results of randomly adding different vehicles near the target parking space.

\begin{table*}[!t]
  \centering
  \caption{Comparative Experiments of Different Parking Methods.}
  \label{tab:comparison}
  \renewcommand{\arraystretch}{1.3}
  \setlength{\tabcolsep}{4pt}
  \begin{tabularx}{\textwidth}{c|c|*{5}{Y}|*{5}{Y}}
    \toprule
    \multirow{2}{*}{\textbf{Type}} & \multirow{2}{*}{\textbf{Methods}} & \multicolumn{5}{c|}{\textbf{Standard Slot}} & \multicolumn{5}{c}{\textbf{Mechanical Slot}} \\
    \cmidrule(lr){3-7} \cmidrule(lr){8-12}
    & & \makecell{\textbf{PSR}\\(\%) $\uparrow$} & \makecell{\textbf{PCR}\\(\%) $\downarrow$} & \makecell{\textbf{PTR}\\(\%) $\downarrow$} & \makecell{\textbf{PBR}\\(\%) $\downarrow$} & \makecell{\textbf{NGS}\\$\downarrow$} & \makecell{\textbf{PSR}\\(\%) $\uparrow$} & \makecell{\textbf{PCR}\\(\%) $\downarrow$} & \makecell{\textbf{PTR}\\(\%) $\downarrow$} & \makecell{\textbf{PBR}\\(\%) $\downarrow$} & \makecell{\textbf{NGS}\\$\downarrow$} \\
    \midrule
    \multirow{3}{*}{\makecell{Rule-based\\Planner}}
    & RS Curve \cite{reeds1990optimal}   & 37.0  & 0.0  & 63.0  & 0.0  & 1.1   & 0.0   & 0.0  & 100.0  & 0.0  & 0.0  \\
    & Hybrid A* \cite{Kurzer1057261}  & 67.5  & 0.0  & 32.5  & 0.0  & 12.1  & 50.0  & 0.0  & 50.0  & 0.0  & 13.1 \\
    \midrule
    \multirow{6}{*}{\makecell{Pretrained \\ Perception \\ + \\RL-based\\Planner}}
    & PPO \cite{schulman2017proximal}  & 42.5  & 33.0  &  17.5  & 7.0   & 29.3 & 37.0  & 47.5  & 10.5  & 5.0 & 30.5  \\
    & SAC \cite{haarnoja2018soft}  & 73.0 & 22.0 & 5.0 & 0.0  & 14.3  & 58.0  & 37.5  & 3.5 & 1.0   & 10.6  \\
    & HOPE \cite{jiang2025hopereinforcementlearningbasedhybrid} & 58.0 & 33.0 & 8.5 & 0.5 & 29.7 & 36.0 & 55.5 & 7.5 & 1.0 & 28.7 \\
    & HOPE \cite{jiang2025hopereinforcementlearningbasedhybrid}  + RS & 72.5 & 26.0 & 1.5 & 0.0 & 3.5 & 48.0 & 51.5 & 0.5 & 0.0 & 2.8 \\
    & RL-OGM-Parking \cite{wang2025rlogmparkinglidarogmbasedhybrid} & 56.5 & 32.5 & 8.5 & 2.5 & 17.4 & 39.0 & 55.5 & 5.5 & 0.0 & 14.3\\
    & SEG-Parking \cite{li2024segparking} & 61.5 & 30.5 & 5.5 & 2.5 & 13.7 & 42.5 & 54.0 & 3.0 & 0.5 & 6.8 \\
    \midrule
    \multirow{4}{*}{\makecell{End-to-end}}
    & ParkingE2E \cite{li2024parkinge2e}  & 44.0  & 38.5 & 0.0  & 17.5 & 1.9   & 38.5  & 50.5 & 0.0  & 11.0 & 3.5  \\
    & REAP-PPO              & 51.5  & 34.5 & 11.0 & 3.0  & 24.0  & 40.5  & 39.0 & 18.5 & 2.0  & 31.6 \\
    & REAP-SAC (Ours)       & \textbf{83.5} & 11.5 & 5.0 & 0.0 & 15.9 & \textbf{70.0} & 28.5 & 1.5 & 0.0 & 9.2 \\
    \bottomrule
  \end{tabularx}\par
  \vspace*{1.2em}
\noindent\begin{minipage}{0.98\textwidth}
\footnotesize \textit{Notes:}\ 
(1) Rule-based methods use a pre-built map. The timeout time of planner is 5 s, and timeout and failures are counted in PTR. 
(2) RL-based planners use BEV input. For fair evaluation, the same pretrained perception model as REAP is used. For the standard staged RL baselines PPO and SAC, the policy follows the same staged training strategy as HOPE.
(3) For HOPE + RS, many repeated gear-switching maneuvers used for garage alignment are directly avoided by the RS guidance at the terminal stage, leading to a lower NGS. However, since HOPE here uses perception results instead of ground truth, the reduced obstacle prediction accuracy leads to a relatively high collision rate. In contrast, REAP does not require RS during inference and still achieves high parking accuracy.
(4) Adapted baseline results are reported under a unified surround-view-camera closed-loop protocol for controlled comparison, rather than as exact reproductions under each method's original sensing and deployment setting.
\end{minipage}\par
\vspace*{-0.9em}
\end{table*}

The perception model adopts BEV architecture, with a perception range of $[-10 \text{m}, 10\text{m}]$ and the resolution of the BEV grid is $0.05 \text{m}$.
The model was trained with an NVIDIA RTX 4090.
The batch size was set to 32.
The model was trained for a total of 20,000 epochs.
For motion prediction, the model predicted actions over a 0.5-second horizon.
In the soft collision reward calculation, the threshold of truncated Euclidean distance $d_0$ is set to $1.0\text{m}$, and the non-linear factor $\tau$ is set to $0.5$.

\textbf{Model Inference and Deployment}: Unlike in training, the target parking space in real-world parking scenarios must be obtained through human–machine interaction. 
We use IPM (Inverse Perspective Mapping) images generated from surround-view cameras to select the target, which is then matched with the parking space detected by the perception module. 
Due to sources of error such as IPM distortion, randomness in target-point selection, and odometry drift, we cannot assume that the target slot location is exact. Therefore, target perturbations are continuously applied during training.
During inference and real-vehicle deployment, the target parking space is continuously tracked using odometry. 

Although the simulator leverages 3DGS technology and vehicle kinematic models to closely replicate the visual inputs and vehicle motion characteristics of the real-world environment, the underlying dynamic control system remains difficult to model accurately. 
To address this, during real-vehicle operation, action commands are converted into trajectories over a fixed time interval and executed by the vehicle controller. 
From a system perspective, the current deployment combines learned safety shaping with controller-level execution: the soft/predictive collision penalties and 2.5D collision-aware simulation improve the policy's safety margin during training, while the vehicle controller executes short-horizon trajectories rather than raw actions during deployment. These measures improve practical safety, although they do not yet constitute a formal safety guarantee.
Real-vehicle testing results demonstrate that our proposed Real2Sim2Real simulator and end-to-end algorithm effectively narrow the gap between simulation and reality.
A vehicle equipped with an onboard computing unit featuring an NVIDIA RTX 2060 GPU was utilized for real-world validation.

\textbf{Experimental Evaluation Metrics}: 
Our evaluation metrics include Parking Success Rate (PSR), Parking Collision Rate (PCR), Parking Timeout Rate (PTR), Parking Boundary-crossing Rate (PBR), Number of Gear Shifts (NGS), and Average Parking Time (APT).
PSR refers to the proportion of collision-free and safe arrivals at the parking space. 


\subsection{Comparative Experiments}

\begin{table*}[!t]
  \centering
  \caption{Ablation Study on Different Modules and Training Strategies of the Algorithm}
  \label{tab:ablation}
  \renewcommand{\arraystretch}{1.3}
  \setlength{\tabcolsep}{8pt} 

  \begin{tabularx}{\textwidth}{l | l | *{5}{Y}}  
    \toprule
    \textbf{} & \textbf{Ablation Term} & \makecell{\textbf{PSR}\\(\%) \(\uparrow\)} & \makecell{\textbf{PCR}\\(\%) \(\downarrow\)} & \makecell{\textbf{PTR}\\(\%) \(\downarrow\)} & \makecell{\textbf{PBR}\\(\%) \(\downarrow\)} & \makecell{\textbf{NGS}\\\(\downarrow\)} \\
    \midrule

    \multirow{1}{*}{}
    \textbf{Baseline} &  retain all strategies   & \textbf{83.5}  & 11.5  & 5.0  & 0.0  & 15.9 \\
    \midrule

    \multirow{2}{*}{\textbf{Training Stage}}
    & w/o perception pretrain (stage 1)           & 50.5  & 38.0  & 10.5  & 1.0  & 19.6  \\
    & w/o hybrid reinforcement learning (stage 2) & 0.0  & 9.5  &  88.5 & 2.0  &  59.8 \\
    \midrule

    \multirow{2}{*}{\textbf{Network Structure}}
    & w/o auxiliary tasks (actor)   &  23.0 &  71.5 &  5.5 & 0.0  & 14.3  \\
    & w/o auxiliary tasks (critic)  & 40.5  & 55.5  & 4.0  & 0.0  & 8.3  \\
    \midrule

    \multirow{2}{*}{\textbf{Reward Design}}
    & w/o parking anchor reward               & 76.5  & 20.0  & 3.5  & 0.0 & 12.3  \\
    & w/o soft collision reward  & 73.5  & 16.5  & 7.0  & 3.0  & 13.7  \\
    & w/o predictive collision reward  & 77.0  & 15.0  & 6.0  & 2.0  & 16.6  \\
    \midrule

    \multirow{2}{*}{\textbf{Perturbation Strategy}}
    & w/o target disturb    &  66.5 & 25.0  & 8.5  &  0.0 & 16.2  \\
    & w/o camera disturb    &  70.0 &  27.0 & 1.5  & 1.5  & 8.0  \\
    \bottomrule
  \end{tabularx}
  \vspace{5pt}
\end{table*}

Table~\ref{tab:comparison} compares REAP with three groups of baselines: rule-based planners (RS Curve and Hybrid A*), RL-based planners (PPO, SAC, HOPE \cite{jiang2025hopereinforcementlearningbasedhybrid}, RL-OGM-Parking \cite{wang2025rlogmparkinglidarogmbasedhybrid}, and SEG-Parking \cite{li2024segparking}), and end-to-end methods (ParkingE2E \cite{li2024parkinge2e}, and REAP).
This taxonomy evaluates whether REAP improves over geometric planners, how it compares with representative RL-based parking methods, and whether SAC brings a clear benefit within the same camera-based end-to-end framework.
Accordingly, we also report REAP-PPO, which uses the same input, network architecture, and training strategy as REAP-SAC.

For fair comparison, each method is evaluated on 200 randomly sampled start-target pairs, and standard and mechanical slots are reported separately because their geometry and parking difficulty differ substantially.
For representative baselines whose original sensing inputs or task formulations differ from ours, we adopt a unified surround-view-camera closed-loop protocol whenever possible.
In particular, non-end-to-end learning baselines are driven by the same pretrained perception outputs used by REAP, making the comparison closer to a perception-in-the-loop deployment setting while keeping the adaptation boundary explicit.
Under this protocol, HOPE computes its action mask from predicted occupancy maps, SEG-Parking uses the same RS/expert dataset for offline RL, and PPO/SAC are trained in the same staged BEV-planner setting.
Only methods that can be adapted to this protocol without changing their core policy structure are included in the direct quantitative comparison.
Methods that would require different state/action formulations or additional search/MPC/federated components are discussed for positioning but not numerically compared.

\begin{figure*}[t]
		\centering
		\includegraphics[width=0.9\textwidth]{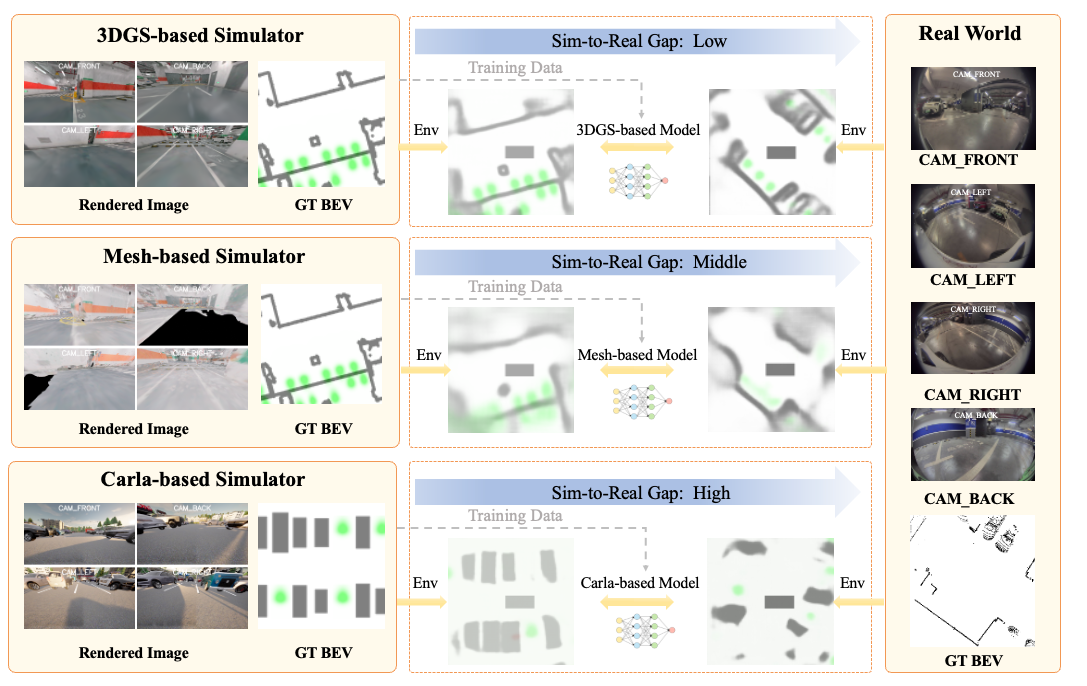}
		\caption{\textbf{Qualitative Sim2Real perception comparison of models trained with different simulators.}
		The three rows in the left block show the rendered environment images and occupancy maps generated by three different simulators: 3DGS-based, Mesh-based, and CARLA-based.
		The right block shows the environment images and occupancy map of the corresponding real-world scene.
		The middle block presents the perception results of models trained with data generated from the corresponding simulator.
		The 3DGS-trained model produces predictions that are more consistent with the real-world occupancy reference, whereas the Mesh- and CARLA-trained models exhibit more noticeable perception degradation, indicating a larger Sim2Real gap.}
		\label{fig:sim2real_perception}
\end{figure*}

REAP-SAC achieves the highest parking success rate in both slot types.
Rule-based methods remain collision-free but often fail in complex multi-step maneuvers; RL-based planners improve exploration but are more sensitive to uncertainty in the perceived BEV state; imitation-based end-to-end baselines are more sensitive to expert-data distribution shift.
By training directly in the camera-based closed-loop setting with target perturbations, REAP-SAC provides stronger parking accuracy and collision avoidance.
Together with the module ablation in Table~\ref{tab:ablation}, the simulator ablation in Table~\ref{tab:simulator_ablation} and Fig.~\ref{fig:sim2real_perception}, and the real-vehicle validation in Table~\ref{tab:real_world}, these comparisons support the effectiveness of the proposed learning framework and Real2Sim2Real pipeline.

\begin{table}[t]
\centering
\begin{minipage}{0.98\linewidth}
\centering
\caption{Ablation Study on Different Simulator Types.}
\label{tab:simulator_ablation}
\scriptsize
\renewcommand{\arraystretch}{1.15}
\setlength{\tabcolsep}{2pt}
\begin{tabular}{l|cccc}
\toprule
\makecell{\textbf{Simulator Type}\\} & \makecell{\textbf{Build Time}\\\textbf{(h) $\downarrow$}} & \makecell{\textbf{FPS}\\\textbf{$\uparrow$}} & \makecell{\textbf{Sim PSR}\\\textbf{(\%) $\uparrow$}} & \makecell{\textbf{Real PSR}\\\textbf{(\%) $\uparrow$}} \\
\midrule
CAD (CARLA) & $>10$ & 130 & 79.5 & 0.0 \\
Mesh & $\approx 1.0$ & 95 & 76.0 & 45.0 \\
3DGS (Ours) & $\approx 1.0$ & 53 & 82.5 & 70.0 \\
\bottomrule
\end{tabular}
\par\vspace{0.9em}
\begin{minipage}{0.88\linewidth}
\footnotesize \textit{Note:} FPS is measured for simulator rendering at a single-image resolution of $480 \times 270$ on a single NVIDIA RTX 4090 GPU. Real PSR is measured on a 20-trial standard-slot real benchmark composed of five trials from each of the four standard parking scenarios.
\end{minipage}
\end{minipage}
\end{table}
\vspace{-0.3em}

\subsection{Module Ablation}
We conducted a comprehensive ablation study of the proposed method.
The ablation study results are presented in Table \ref{tab:ablation}.
The training stage ablation primarily investigates the impact of different training phases on performance.
The results show that each stage has a significant impact on model performance.
The network architecture ablation examines the effect of adding auxiliary tasks on model performance, revealing that supervision from auxiliary tasks can enhance performance.
The reward design ablation demonstrates that the parking anchor reward ensures rapid convergence, while the predictive collision reward and the soft collision reward play distinct but complementary safety roles.
The perturbation strategy ablation investigates the influence of perturbations on model performance.
The perturbation strategy is crucial for ensuring that performance does not degrade when transferring from the simulator to real vehicles.




	\begin{figure*}[t]
		\centering
		\includegraphics[width=0.8\linewidth]{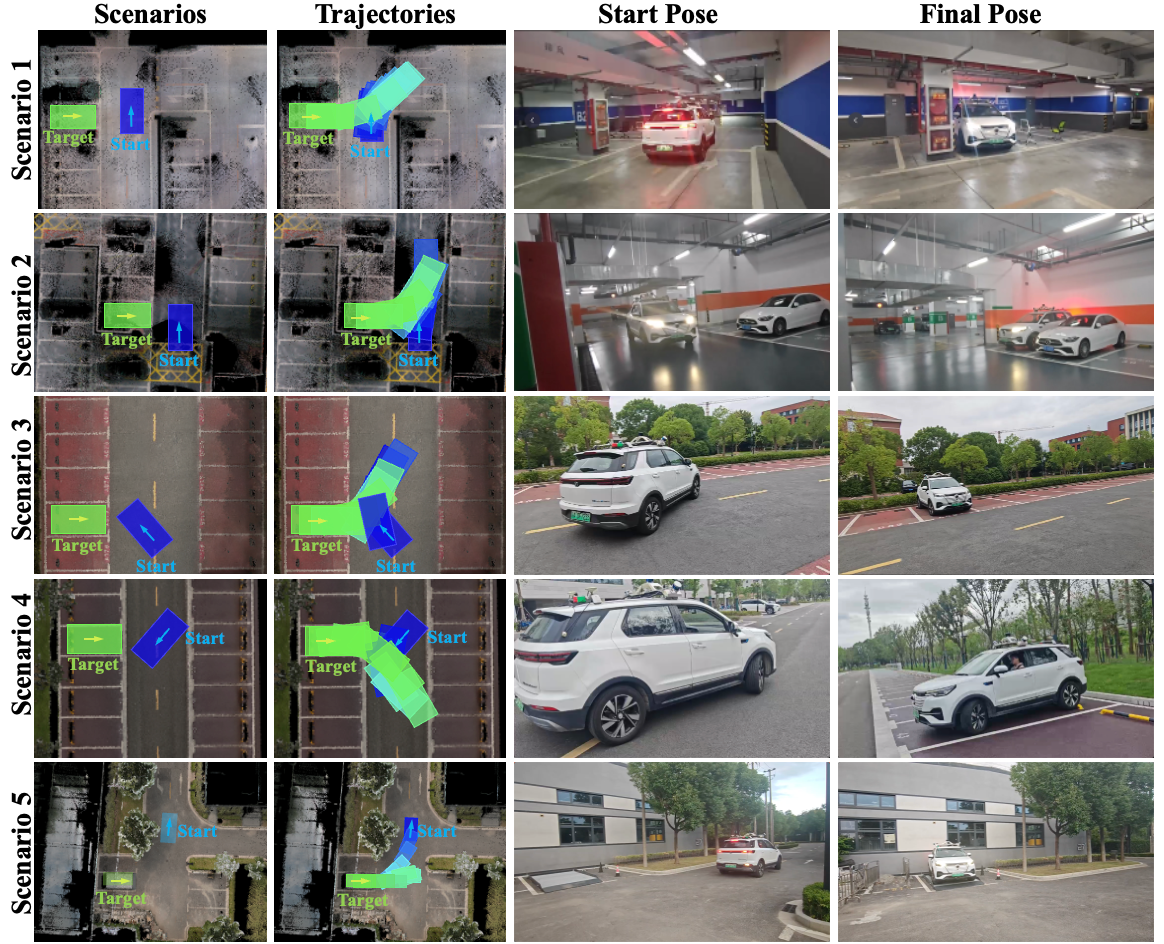}
		\caption{
			\textbf{Representative closed-loop real-vehicle parking results in five test scenarios.} Each row corresponds to one real-world test scenario. From left to right, the four columns show the BEV scene layout, the executed parking trajectory, the real-vehicle start pose, and the final parked pose, respectively. In the BEV scene layout, the blue arrow indicates the initial vehicle pose. The five scenarios cover diverse real parking environments, including underground garages with nearby obstacles, ground parking lots, and a highly constrained narrow parking case, illustrating the Sim2Real capability of the proposed method.
		}
		\label{fig:real_exp}
	\end{figure*}

\subsection{Simulator Ablation}
To quantitatively assess the role of the 3DGS-based simulator and address concerns regarding its computational overhead, we conduct a controlled ablation comparing three categories of visual simulators: 
(1) CAD-based simulator (CARLA \cite{dosovitskiy2017carla}), which relies on manually designed 3D assets; 
(2) Mesh-based simulator, which reconstructs scene geometry from the collected LiDAR point clouds and camera images using traditional mesh reconstruction; 
(3) 3DGS-based simulator (ours), which uses the same collected data but represents the scene with 3D Gaussian Splatting for photorealistic rendering. 
The Mesh-based and 3DGS-based simulators are constructed from identical raw sensor data. To ensure a fair comparison, we further integrate the Mesh-based and CARLA-based rendering engines into our simulator framework, while keeping the data-collection pipeline, vehicle motion model, camera configuration, control interface, and training/evaluation protocol as consistent as possible. Therefore, the primary difference among the compared variants lies in the scene representation and rendering engine itself. Although CARLA is primarily designed for urban driving, it includes a subset of garage-like scenes, making this parking-oriented comparison feasible. 

As shown in Table~\ref{tab:simulator_ablation}, the CAD-based simulator requires substantial manual effort to construct parking environments, whereas the Mesh-based and 3DGS-based approaches leverage the same scanned data and can be built within minutes. 
Concretely, a standard underground parking lot with around 100 parking spaces can be scanned and digitized in 10 to 15 minutes, and the standalone 3DGS renderer exceeds 50 FPS at a resolution of $480 \times 270$ on a single NVIDIA RTX 4090 GPU. This is consistent with the 53 FPS reported in Table~\ref{tab:simulator_ablation}, since the latter reflects the throughput of the compared simulator setup rather than the peak speed of the standalone renderer.
Most importantly, the 3DGS-based simulator yields the highest Sim2Real transfer performance: models trained in the 3DGS environment achieve substantially higher real-vehicle parking success rates, confirming that photorealistic visual fidelity is critical for bridging the domain gap.
Fig.~\ref{fig:sim2real_perception} further illustrates this advantage qualitatively: the perception module trained in the 3DGS simulator produces accurate and consistent outputs when deployed on the real vehicle, whereas models trained with lower-fidelity rendering exhibit noticeable perception degradation.

\subsection{Real-world Experiments}

\begin{table}[t]
\centering
\caption{Real-Vehicle Experimental Results.}
\label{tab:real_world}
\setlength{\tabcolsep}{5.0pt}
\begin{tabular}{c|ccccc}
\toprule
\textbf{Scenarios} & \makecell{\textbf{PSR}\\(\%) \(\uparrow\)} & \makecell{\textbf{PCR}\\(\%) \(\downarrow\)} & \makecell{\textbf{PTR}\\(\%) \(\downarrow\)} & \makecell{\textbf{PBR}\\(\%) \(\downarrow\)} & \makecell{\textbf{APT}\\(s) \(\downarrow\)} \\
\midrule
Scenarios 1 & 65.0 & 25.0 & 10.0 & 0.0 & 77 \\
Scenarios 2 & 70.0 & 20.0 & 5.0 & 5.0 & 71 \\
Scenarios 3 & 85.0 & 5.0 & 0.0 & 10.0 & 57 \\
Scenarios 4 & 80.0 & 10.0 & 5.0 & 5.0 & 64 \\
Scenarios 5 & 55.0 & 25.0 & 20.0 & 0.0 & 89 \\
\bottomrule
\end{tabular}

\end{table}

To validate the Sim2Real transfer capability of the algorithm, we conducted closed-loop real-vehicle experiments across five distinct scenarios. 
In each scenario, 20 parking trials with randomly selected start-target configurations were used for testing, and the results are reported in Table~\ref{tab:real_world}.
	Scenarios 1 and 2 correspond to underground parking garages with nearby obstacles, Scenarios 3 and 4 correspond to ground parking lots with different initial poses, and Scenario 5 represents the most challenging narrow parking case.

As shown in Table~\ref{tab:real_world}, the proposed method achieves a Parking Success Rate (PSR) of $65\%$--$85\%$ across the four standard parking scenarios, with the best performance of $85\%$ PSR observed in Scenario 3. 
In the most challenging case—the mechanical parking slot (Scenario 5)—the algorithm still achieves a PSR of $55\%$ with an average parking time of $89$ seconds, demonstrating the effectiveness of the proposed 2.5D collision simulation and the hybrid training strategy for this extreme task.

Comparing the real-vehicle results with the simulation performance in Table~\ref{tab:comparison} (where REAP-SAC achieves $83.5\%$ PSR on standard slots and $70.0\%$ on mechanical slots), the real-world results are broadly consistent. 
This close correspondence between simulation and real-world performance provides quantitative evidence that the 3DGS-based simulator substantially narrows the domain gap, enabling effective Sim2Real transfer. 
The primary source of performance degradation in the real world is the dynamic control error, which is not fully modeled in the simulator.

Fig.~\ref{fig:real_exp} presents representative real-vehicle results for these five scenarios; each row shows the BEV scene layout, executed trajectory, start pose, and final pose, making the parking process easier to interpret visually.

\section{Limitations and Conclusion}
This paper proposes a reinforcement learning-based end-to-end parking algorithm that leverages rule-based planner to address initial convergence challenges and employs soft predictive collision penalties to effectively reduce collision rates. The algorithm, trained in the 3D Gaussian splatting simulator and deployed on real-world parking spaces (standard parking space and narrow mechanical parking slot),  demonstrates effective Sim2Real transfer capability and promising parking performance.
To the best of our knowledge, we are among the first to validate a parking algorithm using a Real2Sim2Real vehicle navigation simulator and successfully transfer it to real-vehicle deployment, while accomplishing the task effectively with minimal data collection cost and training resources.

\textbf{Limitations}: Despite the strong Sim2Real performance, several limitations remain. 
First, reinforcement learning fine-tuning improves success rates but increases gear shifting, slightly reducing parking smoothness. 
Second, the current training pipeline still benefits from planner-derived supervision: rule-based actions stabilize early-stage hybrid training, and the parking anchor reward is defined by states with feasible rule-based trajectories. The final deployed policy does not require planner queries during inference, but reducing this training-time dependence and enabling stronger autonomous exploration remain important future directions. 
Third, although our experiments cover underground garages, ground parking lots, and extremely narrow mechanical parking slots, the validation is still focused on relatively structured scenarios. Generalization to unseen outdoor environments and highly dynamic obstacles therefore requires further study. 
Finally, deploying end-to-end models on onboard edge devices still introduces noticeable computational overhead. Reducing latency and resource consumption remains important for lighter automotive hardware. Moreover, the current safety design improves empirical safety but does not provide formal guarantees; integrating deterministic safety filters or hard-constrained fallback planners will be important for broader deployment.


{
    \small
    \bibliographystyle{IEEEtran}
    \bibliography{ref}
}



\end{document}